# Constructing Situation Specific Belief Networks

**Suzanne M. Mahoney**
Information Extraction and Transport, Inc.
1730 N. Lynn Street, Suite 502
Arlington, VA 22009
suzanne@iet.com

**Kathryn Blackmond Laskey**
Dept. of Systems Engineering
George Mason University
Fairfax, VA 22032-4444
klaskey@gmu.edu

## Abstract

This paper describes a process for constructing situation-specific belief networks from a knowledge base of network fragments. A situation-specific network is a minimal query-complete network constructed from a knowledge base in response to a query for the probability distribution on a set of target variables given evidence and context variables. We present definitions of query completeness and situation-specific networks. We describe conditions on the knowledge base that guarantee query completeness. The relationship of our work to earlier work on KBMC is also discussed.

## 1 INTRODUCTION

The first applications of belief networks were in domains that could be characterized by a relatively small and fixed number of variables, for which it was feasible to construct a fixed model in advance of problem solving. In more complex domains it is necessary to reason about a variable number of entities that may be related to each other in varied ways. It is also necessary to reason about and distinguish between multiple instances of a given complex pattern of entities and relationships. In such domains it is infeasible to construct a complete belief network encompassing all the situations one might encounter in problem solving. Knowledge-Based Model Construction (KBMC) is the process of constructing a model for a problem instance from a knowledge base representing generic domain entities and their interrelationships. A KBMC system includes a knowledge base, search operators for retrieving problem-relevant knowledge base elements, network construction operators, network evaluation operators, and model construction control mechanisms. Objectives for a KBMC system are to minimize costs of representation, retrieval, construction and evaluation, while providing accurate responses to queries.

Most research in KBMC has focused on bottom-up model construction, in which a model is built up incrementally until a response to the query can be computed (Breese, 1990; Goldman and Charniak, 1993; Provan, 1993; Haddawy, 1994; Glesner and Koller, 1995). Another line of research concerns top-down model reduction, in which one prunes parts of a large model to obtain a smaller model from which the query response is computed (Lin and Druzdzel, 1997; Baker and Boult, 1991; Geiger et al., 1990; Shachter, 1986). This paper takes the bottom-up construction approach, but draws from results in the literature on top-down reduction.

In this paper we propose a model construction control strategy for producing situation-specific networks from a knowledge base of network fragments (Laskey and Mahoney, 1997). We give conditions on the knowledge base that guarantee completeness and consistency of the implicitly encoded probability model. We then relax the completeness condition and give conditions for the knowledge base to be *query complete* for a given query. A situation-specific network is a minimal (in a sense formally defined below) network sufficient to respond to a query for which the knowledge base is query complete. Our work extends and modifies earlier work on knowledge-based model construction. Our use of the object-oriented paradigm for both knowledge representation and construction provides the ability to represent abstract types with associated structure, methods, and inheritance. Network fragments may represent semantically meaningful chunks of knowledge, or commonly co-occurring parts of a belief network that should be retrieved as a unit. Influence combination provides a parsimonious representation for local structure and a mechanism for hypothesis generation. We make use of nuisance nodes (Lin and Druzdzel, 1997) to simplify constructed networks. When the set of likely queries is known in advance, the knowledge base can be preprocessed in the background to marginalize over nuisance nodes and increase run-time efficiency. Finally, we make use of context nodes to index the distributions retrieved from the knowledge base.

## 2  BACKGROUND

### 2.1  NETWORK FRAGMENTS

Network fragments (Laskey and Mahoney, 1997) provide an object-oriented representation for probabilistic knowledge. A probability model is represented implicitly as a knowledge base of belief network fragments. Each fragment consists of a set of random variables connected by a fragment graph, together with information used to construct local distributions for variables. Variables in a fragment are classified as either resident or input. The information needed to define the distributions for random variables is carried in fragments in which they are resident.

Both fragments and random variables are objects organized in a type hierarchy, with associated structure and methods. Each random variable and fragment has a set of *identifying attributes* which, when specified, create a unique identifier for an instance of the random variable or fragment. There is a mapping from identifying attributes of the fragment to identifying attributes of its random variables. Identifying attributes play the role of variables in a logic programming language (to avoid confusion, we reserve the term variable for random variables).

When a variable's probability distribution has local structure, it is often convenient to specify its distribution in several different fragments to be combined at run time into a full distribution. For example, the "Disease" node in a medical diagnosis system may be specified as a noisy-OR distribution with perhaps hundreds of input diseases. These input distributions could be organized into groups of related diseases, each represented as a separate fragment. At run-time there may be information available that rules out entire categories of diseases, requiring only a small percentage of the groups to be included in the final constructed model.

We use *influence combination* to represent local structure (Laskey and Mahoney, 1997). Random variables have an *influence type* with associated *influence combination method*. Enabling conditions for a given influence type specify restrictions on the influencing variables. For example, a noisy-OR variable must take only binary input variables. Each individual fragment uses an *influence function* to represent parameters of the influence combination method. For example, the trigger probability for a candidate cause in a noisy-OR distribution is returned by the influence function in the fragment where the cause-effect link is defined.

### 2.2  TOP-DOWN NETWORK REDUCTION

Lin and Druzdzel (1997) considered the problem of pruning a large belief network down to a network sufficient to respond to a query specified as

$q = P(\underline{T} \mid \underline{E})$, where T is a set of *target* variables and E is a set of *evidence* variables. Lin and Druzdzel first reduce the network to a set of computationally relevant nodes using a combination of $d$-separation (Geiger, et al, 1990) and barren node removal (Baker and Boult, 1990). Consider the graph of Figure 1, and consider the query $P(\underline{T}|\underline{E})$, where $\underline{T}$ and $\underline{E}$ represent vectors or lists of nodes labeled with the respective letter. [1] In this network, the nodes D1 through D6 can be removed because they are $d$-separated from the target nodes given the evidence nodes. The nodes B1 through B5 are barren (that is, they have no descendants that are either target or evidence nodes). These nodes also can be removed without affecting the result of the query. When these nodes and their associated arcs are removed, the result is a *computationally relevant network*, or the minimal subnetwork from which the response to the query can be computed. An interesting property of the computationally relevant network is that all nodes must be evidence nodes, target nodes, or ancestors of one or more target nodes.

The next step in Lin and Druzdzel's graph reduction is the marginalization and removal of *nuisance nodes.* Lin and Druzdzel define a nuisance node as a node that is computationally relevant given the query, but is on no evidential trail[2] between an evidence and a target node. In Figure 1, the nodes labeled N1 through N5 are nuisance nodes, as are D5 and D6 (which are also $d$-separated from the target nodes given the evidence nodes). Nuisance nodes have the interesting property that the way in which they enter into computation of the query response does not depend on the evidence. Thus, they can be marginalized out and removed prior to declaring and propagating evidence without changing the result of the query.

Nuisance nodes in any belief network can be decomposed into nonoverlapping subgraphs, in which each subgraph contains only ancestors of a single node on an evidential trail. There are five such subgraphs in Figure 1: the nuisance tree containing nodes N1 and N2, the nuisance graph containing nodes N4 and N5, and three single-node nuisance trees containing N3, D5, and D6, respectively. Note that D5 and D6 qualify as both d-separated and nuisance nodes. The single node at the base of a nuisance tree or graph is called a *nuisance anchor*.

Before calculating the response to a query, Lin and Druzdzel marginalize the nuisance nodes into their nuisance anchors. Their experimental results show that the cost to reduce the network is more than offset by the savings in inference cost. If common queries are known in advance, a preprocessor can be run offline to compute and cache distributions for nuisance anchors in which nuisance nodes have been marginalized out. Such preprocessing can result in enormous savings in computation.

---

[1] By convention, we use an underscore to indicate a vector or list. We use uppercase letters to represent random variables. Lowercase letters represent states, or possible values of random variables. Boldface lowercase letters represent identifying attributes, which are used to quantify random variables. Greek letters represent ground symbols that denote specific instances of random variables.

[2] An evidential trail between two sets of nodes is a minimal active undirected path from a node in one set to a node in the other.

# 3  SITUATION SPECIFIC NETWORKS

## 3.1  DEFINITION

Bottom-up network construction incrementally builds a belief network to respond to a query. It is assumed that the knowledge base implicitly encodes a complete probability model that is never explicitly constructed.[3] The knowledge base contains *variables*, objects that represent a random variables. The variable $V(\underline{x})$ refers to a random variable with name $V$ and identifying attributes $\underline{x}$. Dependence between variables is represented by directed arcs in fragment graphs, which we call *links*.

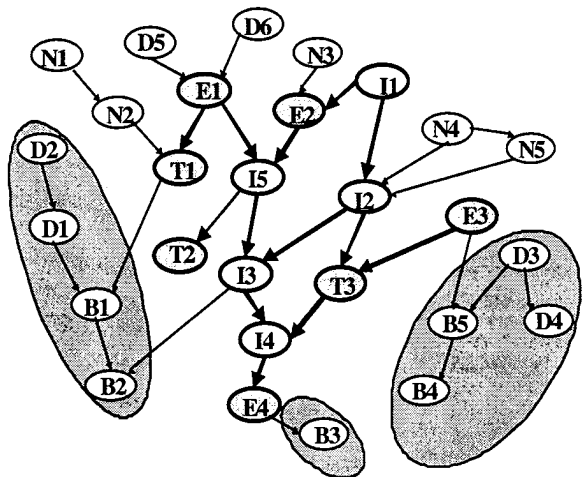

Figure 1 Example Bayesian Network

Network construction is initiated in response to a query

$$Q:\ P(\underline{T}(\underline{\alpha})|\underline{C}(\underline{\beta})=\underline{c}(\underline{\beta}),\underline{E}(\underline{\gamma})=\underline{e}(\underline{\gamma})) \quad (1)$$

requesting the probability distribution for a set of target random variable instances given the states of a set of context random variable instances and a set of evidence random variable instances. The symbols $\underline{\alpha}$, $\underline{\beta}$, and $\underline{\gamma}$ fill the identifying attribute slots of the random variables $\underline{T}(\underline{x})$, $\underline{C}(\underline{y})$, and $\underline{E}(\underline{z})$, respectively. In response to the query, a belief network is constructed in a *model workspace*. The nodes in the constructed network refer to instances of random variables in the knowledge base and the arcs correspond to links in fragment graphs in the knowledge base.

In our framework, context plays a similar role to evidence in that queries are conditional on given values of context variables. Colloquially, the word context is usually used to refer to knowledge that remains fixed over a large class of queries for which evidence and targets may vary. Generally, the use of context serves to simplify inference by restricting attention to relevant portions of the model. We use context to index distributions to be retrieved. We are developing a general framework for reasoning with context. For this paper, we simply assume:

1) The cross-product of the state spaces of the context variables is partitioned into a mutually exclusive and exhaustive set of *contexts*;

2) Each network fragment in the knowledge base points to one or more of these contexts;

3) The conditioning event for any query is consistent with exactly one of these contexts.

Conceptually, context variables can be treated as root variables in the belief network implicitly encoded in the knowledge base. If each query refers to exactly one context, then distributions for context variables need not be defined.

During network construction, creation of random variable instances triggers the retrieval of network fragments in which the corresponding random variables are resident. The identifying attributes of the random variable in a retrieved fragment is unified with the values of the identifying attributes of the instance that triggered its retrieval. These identifying attribute values are also unified with the same identifying attributes appearing in other random variables in the fragment. This process creates new random variable instances, which triggers the retrieval of fragments in which they are resident.

Many KBMC systems are limited to this process of instance creation and retrieval of model components unifying with existing instances (Haddawy, 1994; Breese, 1987). In such systems, the only ground symbols appearing in the constructed network are those in·the evidence nodes for the query. Charniak and Goldman (1991; Goldman and Charniak, 1993) permit domain entities not mentioned in the triggering query to be hypothesized. For example, the mention of an action in a plan may trigger the system to hypothesize an agent to carry out the plan and a location where the plan is carried out, neither of which may be explicitly mentioned in the sentence being processed. We require this ability to hypothesize unmentioned entities. For example, a report of a surface-to-air missile (SAM) site may trigger the system to hypothesize other SAMs in the same battalion, a battalion command post, and a military asset that the site is defending. We include a brief discussion of hypothesis management, but defer a full treatment to a later paper.

---

[3] The full network may be infinite in domains with an indefinite and unbounded number of objects (Goldman and Charniak, 1993; Laskey and Mahoney, 1997).

The goal of network construction is to construct a minimal network, called a *situation-specific network*, sufficient to respond to a query.

**Definition:** Let $Q = P(\underline{T}(\alpha)|\underline{C}(\beta)=\underline{c}(\beta), \underline{E}(\gamma)=\underline{e}(\gamma))$ be a query. Let $N$ be a Bayesian network whose nodes include the random variable instances $\underline{T}(\alpha)$ and $\underline{E}(\gamma)$.[4] $N$ is a *situation-specific network* if all nodes in $N$ are either target nodes or belong to evidential trails between target and evidence nodes.

## 3.2 COMPLETENESS

A situation-specific network is constructed by retrieving fragments from the knowledge base, inserting symbols in place of quantified variables, and merging identical nodes. To ensure that constructing a situation-specific network is possible requires placing constraints on the knowledge base. Haddawy (1994) describes conditions on the knowledge base that guarantee an isomorphic mapping between queries and constructed Bayesian networks. The following conditions are slightly more general than Haddawy's. They are sufficient to guarantee that the knowledge base implicitly encodes a unique probability distribution conditional on each context.

1) Every non-context random variable is resident in at least one fragment given each context.

2) If a random variable is resident in more than one fragment in a given context, the enabling conditions for influence combination are met for the collection of fragments in which it is resident.

3) Identifying attributes for a random variable in a fragment are also identifying attributes for its children. More precisely, the fragment identifying attribute that maps to an identifying attribute for a variable also maps to an identifying attribute for each of its children.

4) The link graph for the knowledge base contains no directed cycles.

Conditions 1 and 2 guarantee that each non-context random variable has a local distribution. In model construction, input variables to instantiated fragments in the model workspace trigger retrieval of fragments in which they are resident. Condition 3 guarantees that all identifying attributes in the final situation-specific network will be assigned values. Condition 4 ensures that no cycles exist in the constructed network.

A knowledge base satisfying conditions 1 through 4 is called *strongly complete*. Strong completeness implies that there is a unique probability distribution implicitly encoded in the knowledge base and that for any query the construction algorithm given below terminates in a response consistent with this distribution.

If Condition 3 is relaxed but 1, 2 and 4 are satisfied, the knowledge base is *complete*. When Condition 3 is violated, fragment retrieval and instantiation may leave identifying attributes unspecified in ancestors of the variable triggering retrieval. These identifying attributes must be specified to create the random variable instances for the situation-specific network. The values of these identifying attributes may be associated with existing instances of their corresponding variables, or new instances with different identifying attributes may be postulated. This process is called hypothesis management, and requires modification of the algorithm below (cf., Goldman and Charniak, 1993; Laskey and Mahoney, 1998).

If Condition 1 is relaxed there may be variables in the knowledge base for which no distribution is specified. If the resulting situation-specific network provides a unique response to a given query, then the knowledge base is *query complete* for that query.

# 4    NETWORK CONSTRUCTION

## 4.1 FRAGMENT RETRIEVAL

The network constructor responds to a query by generating a network in the model workspace using three basic operations: fragment retrieval, variable instantiation, and fragment combination.

A fragment retrieval request specifies a variable, $R(\varrho)$, and a context $\underline{C}(\beta) = \underline{c}(\beta)$. Fragments **F** are retrieved containing the resident variable $R(\underline{x})$ and matching the context. The identifying attribute slots $\underline{x}$ in $R$ are set to values $\varrho$. In addition, the mapping between fragment identifying attributes and variable identifying attributes is used to fill all identifying attribute slots corresponding to $\underline{x}$.[5] After retrieval and variable instantiation, the fragment is merged with existing fragments in the model workspace by the graph union operation and the application of influence combination when $R(\underline{x})$ is resident in multiple fragments.

The context of the fragment must be more general than the context of the request. Therefore, $\underline{C}(\beta)$ must include the context variables for all retrieved fragments. Moreover, if a context variable $D(\beta)=d(\beta)$ appears in the query, then the context pointed to by the fragment must contain the value $d(\underline{x})$.

## 4.2 BOTTOM-UP CONSTRUCTION

To simplify the presentation of the network construction algorithm we assume initially that the model workspace is empty. Only minor modifications are required to extend the algorithm to cover incremental extension of existing models to respond to new queries.

---

[4] Context nodes index distributions to retrieve and need not be included in the constructed network.

[5] It is possible that the retrieved fragment will contain nodes with unspecified identifying attributes. Strong completeness ensures that nodes with unfilled identifying attributes will be barren after bottom-up construction terminates. Our construction algorithm removes them automatically.

Simple bottom-up construction builds a Bayesian network **B** upward from the evidence and target variables of a query, adding nodes recursively until all ancestors of the evidence variables have been added. The knowledge base **KB** of network fragments provides the distributions and links to the parents for all the variables. Construction makes use of the methods defined below. The Bayesian network is constructed in a model workspace **MW**. Methods used in the construction are denoted by italicized function names, with parameters enclosed in parentheses. The method *Method*() associated with object **O** is denoted by **O**.*Method*().

*Instantiate* is a model workspace method for constructing nodes for a Bayesian network. It takes a variable, $U(\underline{x})$ , and its identifying attributes, $\underline{\delta}$ and returns the corresponding node, $U(\underline{\delta})$. The method has two optional parameters: a value $u$ to be declared as evidence, and a local distribution $ld$. Another model workspace method is *NewBayesNet*(), which creates an empty Bayesian network object with no nodes or arcs.

*FindFragment* is a knowledge base method for retrieving network fragments. It takes a variable instance, $U(\underline{\delta})$ and a context vector, $\underline{C_\delta}(\underline{\beta})=\underline{c}(\underline{\beta})$ as parameters and returns a network fragment instance, $\mathbf{F}_{U(\underline{\delta})}$ in which $U(\underline{\delta})$ is a resident variable.

BayesNet methods include: *FindNode* which takes either a variable or a node as a parameter, *AddNode* which takes a node as a parameter and *JoinParent* which takes parent node and child node as parameters and unifies a parent of the local distribution of the child node with the parent node. All three methods return a Boolean value indicating whether the operation was successful.

NetworkFragment methods include *GetLocal-Distribution*, *GetParent*, and *GetIdentifyingAttributes*, all with obvious meanings.

*SimpleBottomUpConstruction* is a model workspace method for Bayesian networks that takes a query, $Q$, and a knowledge base, **KB**, as parameters and returns a computationally relevant Bayesian network, **B**. The method operates as follows:

Given query $Q$: $P(\underline{T}(\underline{\alpha})|\underline{C}(\underline{\beta})=\underline{c}(\underline{\beta}),\underline{E}(\underline{\gamma})=\underline{e}(\underline{\gamma}))$:

1)    Initialize:

◊   **B** = **MW**.*NewBayesNet*()

◊   S=∅, where S is a list of parent/child sets waiting to be added to the network

2)    For each $C(\underline{\beta}) = c(\underline{\beta})$, **B**.*AddNode*(*Instantiate*(*C*, $\underline{\beta}$, $c(\underline{\beta})$, ∅))

3)    For each target and evidence node $U(\underline{\delta})$ mentioned in the query:

◊   Add < $\{U(\underline{\delta}), \underline{C}(\underline{\beta})=\underline{c}(\underline{\beta})\}$, ∅> to $S$

4)    Until S is empty, remove <$\{U(\delta), \underline{C}(\underline{\beta})=\underline{c}(\underline{\beta})\}$, V(ε)>, from $S$

◊   If not **B**.*FindNode*($U(\underline{\delta})$):

- $\mathbf{F}_{U(\underline{\delta})}$ = **KB**.*FindFragment*($U(\underline{\delta})$, $\underline{C}(\underline{\beta})=\underline{c}(\underline{\beta})$).

- $ld$ = $\mathbf{F}_{U(\underline{\delta})}$.*GetLocalDistribution*($U(\underline{\delta})$)

- $U(\underline{\delta})$ = *Instantiate*($U$, $\underline{\delta}$, $u$, $ld$).

- **B**.*AddNode*($U(\delta)$).

- For each $P = \mathbf{F}_{U(\underline{\delta})}$.*GetParent*($U(\delta)$)

- Create $\{P(\underline{\delta}_P)$, $\underline{C}(\underline{\beta})=\underline{c}(\underline{\beta})\}$, where $\underline{\delta}_P$ = $\mathbf{F}_{U(\underline{\delta})}$.*IdentifyingAttributes*($P$)

- Add <$\{P(\underline{\delta}_P)$, $\underline{C}(\underline{\beta})=\underline{c}(\underline{\beta})\}$,$U(\underline{\delta})$> to S.

◊   **B**.*JoinParent*($U(\underline{\delta})$,$V(\underline{\varepsilon})$)

5)    *Return* **B**.

Consider the example modeled in Figure 1. The variables with the gray ellipses behind them are those that are not constructed by simple bottom-up construction. They include nodes that are d-separated from the target variables by the evidence variables; D1, D2, D3 and D4: and nodes normally eliminated by barren node removal; B1, B2, B3, B4 and B5. Note that some d-separated nodes remain in the constructed model; D5 and D6.

Simple bottom up construction produces a network containing all computationally relevant nodes using only local knowledge about a variable's parents to guide search. Many network construction systems are based on variants of bottom up construction (Haddawy, 1994; Ngo, et al, 1996; Breese, 1991). Given unlimited resources and a complete knowledge base, the constructed network obtains the same query response as the complete network.

### 4.3 SITUATION-SPECIFIC NETWORKS

Although a situation-specific network can be obtained from the computationally relevant network by removing nuisances nodes (Lin and Druzdzel, 1997), search and computation can be reduced by direct generation of a situation-specific network. This requires a means of identifying nuisance nodes for a query, as well as marginal distributions for nuisance anchors.

Nuisance node detection (Lin and Druzdzel, 1997) needs the graph of all of the ancestors of a group of target and evidence variables to determine which ancestors are nuisance nodes. This information, although nonlocal, requires only the graph of the computationally relevant network. We therefore build the *query graph* first and use it to guide the construction of the situation-specific network.

Experience has shown that structural relationships are relatively static once they have been elicited. Therefore, storing graphs containing nonlocal information is likely to be relatively maintenance free. We use three strategies in combination for obtaining query graphs:

1) Store query graphs for common queries;

2) Store ancestor graphs for likely evidence and target variables. Take the graph union of the ancestor graphs for the target and evidence variables to form a predecessor graph. Then eliminate the nuisance nodes.

3) Use simple bottom-up construction to construct the predecessor graph. Then eliminate the nuisance nodes.

An *ancestor graph* is the graph of a variable's ancestors. For each vertex in the graph, it specifies the variable name and identifying attribute type. An ancestor graph object has a method for consistently assigning a set of identifying attribute values to the vertices in the graph. The graph only includes ancestors whose identifying attribute values are a subset of those specified for the variable. Restricting the identifying attributes values to those of the variable eliminates the infinite regression backward in time or through space via temporal and spatial dependencies.

A *predecessor graph* is the graph union of a set of ancestor graphs whose identifying attribute values have been set. Each vertex corresponding to an evidence or context variable is marked as evidence. Each vertex corresponding to a target variable is marked as target.

A practical strategy is to implement all three approaches. Query graphs for common queries can be easily stored with the network fragments for target and/or evidence variables. A more general approach is needed for less common queries. The ancestor graph alternative allows one to flexibly construct a query graph for any arbitrary query given that each target and evidence variable's fragment caches an ancestor graph for the variable. Finally, the third approach always works if the knowledge base is query complete, although it is the most expensive approach.

As Lin and Druzdzel (1997) have shown, caching alternative conditional probability distributions for a nuisance anchor is beneficial. The obvious place to cache alternative distributions is with the network fragment in which the nuisance anchor is resident. In conjunction with the influence functions for each parent variable, the influence combination method for a nuisance anchor takes into account the parents that are to be explicitly represented in the network. For a noisy-OR distribution, the leak probability may depend upon which parents are explicitly represented. For conditional probability tables, each possible combination of explicitly represented parents may have its own table. In general, the alternative distributions for a nuisance anchor may be obtained by constructing the nuisance nodes, conditioning on the remaining parents and performing inference. This approach may be performed in the background or during network construction. Once computed, results may be cached and reused.

Maintaining query graphs, ancestor graphs and alternative distributions requires that a network fragment be notified when an ancestor fragment is modified. While ancestor graphs and alternative distributions may be used to efficiently construct situation-specific networks, the cost of maintenance is relatively high in knowledge bases that are regularly updated. For static portions of the knowledge base, the cost is mostly the storage cost.

In summary, constructing a situation-specific network from a set of fragments has the following steps: (1) construct a predecessor graph from stored ancestor graphs; (2) reduce the predecessor graph to the query graph by eliminating vertices of nuisance nodes; (3) using the query graph as a guide, instantiate the network fragments.

## 4.4    EXAMPLE

We consider an example from the domain of military situation assessment. The fragments in our knowledge base contains represent the behavior of military units given environmental, mission, equipment and organizational constraints.

**Example query:**

*Unit A is an artillery unit and its echelon is battalion. We receive a report that unit A is at location X at time $t_1$. The report also states that A is moving. We believe*

| Variable Type | Variable Name | Identifying Attributes | State |
|---|---|---|---|
| Target | Activity | Unit A, Time $t_2$ | |
| Target | Location Suitability | Location Y, Unit B, Time $t_1$ | |
| Target | Location Suitability | Location Z, Unit B, Time $t_1$ | |
| Context | Unit Type | Unit A | Artillery |
| Context | Echelon | Unit A | Battalion |
| Context | Unit Type | Unit B | Artillery |
| Context | Echelon | Unit B | Battery |
| Context | Weather | | Hard Rain |
| Evidence | Location Report | Location X, Unit A, Time $t_1$ | Seen Here |
| Evidence | Movement Report | Unit A, Time $t_1$ | Moving |
| Evidence | Communications Report | Unit B, Time $t_1$ | Not Moving |
| Evidence | Unit-Unit Reporting | 1-Unit A, 2-Unit B | 2 is part of 1 |
| Evidence | Terrain | Location Y | Flat w. cover |
| Evidence | Terrain | Location Z | Flat and open |

Table 1 Query Specification

that there is a second military unit B operating in the same area. A communications report intercepted at time $t_1$ indicates that unit B is probably stationary. If B is a battery in A's artillery battalion, how suitable are locations Y and Z for B? Given that it is raining hard, will A still be moving at time $t_2$?

Table 1 summarizes the query to the network constructor. It includes sets of target variables whose states are unknown and sets of context and evidence variables whose states are known. It specifies the identifying attributes including their values for all of the variables.

Figure 4 shows some of the network fragments that will be used to construct the situation specific network for this query. Each fragment is identified by capital letters in its upper left-hand corner. For example, fragment LS has two context variables: "Unit Type = Artillery" and "Echelon = Battery." They are shown with dark gray backgrounds. These variables serve to index the fragment as well as to influence variables that are resident in the fragment. For example, the "Location Suitability" variable resides in other identically structured network fragments in which "Unit Type" and "Echelon" have different values. A variable with a light gray background is a possible evidence variable. "Terrain" is an evidence variable. The fragment LS specifies the distributions for "Location Suitability" and "Terrain" given the specified context variables' values and the "Activity" variable. The "Activity" variable, shown with a black background, is an input variable to the fragment. Only its states are specified in the fragment. The types of identifying attributes for each of the variables are shown below the variable name. Note that the identifying attributes for a child variable contain all of the identifying attributes for the parent variables.

We first construct the predecessor graph.

*We know that Unit A is an artillery unit and its echelon is battalion.* The constructor retrieves fragments for "Unit Type" and "Echelon," in response to *FindFragment* requests. The "Unit" identifying attribute in each fragment is unified with the value A, and the nodes are marked as evidence.

*We receive a report that Unit A is at location X at time $t_1$.* The constructor retrieves fragment LR from the knowledge base in response to *FindFragment*. Its parameters are the variable "Location

Report"(Unit A, Location X, Time $t_1$), and the context "Unit Type"(Unit A)=Artillery and "Echelon"(Unit A) = Battalion. The fragment is retrieved because: evidence variable "Location Report" is resident; the fragment identifying attributes, Location, Unit and Time, are a subset of those specified in the request; and the fragment context is more general than that of the request. The constructor obtains a copy of the ancestor graph of "Location Report" from LR and unifies the identifying attributes for "Location Report" with the corresponding identifying attributes for the ancestor graph. See Graph A in Figure 2. The ancestor graph includes an "Activity" variable only for $t_1$ because only a single time is associated with a given "Location Report". The node "Location Report" is marked as evidence.

*The report also states that A is moving. An intercepted communications report at time $t_1$ indicates that Unit B is probably stationary. If B is a battery in A's artillery battalion, how suitable are locations Y and Z for B?* Following the pattern of fragment retrieval succeeded by graph manipulation illustrated for "Location Report," the constructor retrieves Fragments M, C, R, and LS. It retrieves the ancestor graphs and unifies them with the values of the identifying attributes. In the case of LS, two copies of the ancestor graph are made, one for Location Y and another for Location Z. Some of the resulting graphs are shown as Graphs B and C in Figure 2.

*Given that it is raining hard, will A still be moving at time $t_2$?* This time the retrieval request is *FindFragment* for "Activity"(Unit A, Time $t_1$, Time $t_2$), with context and evidence nodes. The constructor proceeds in reverse order through an ordering for the query graph. This is because fragments for the leaf nodes are always required and a fragment may have several of the variables that need to be constructed. For example, see fragment M in Figure 4.

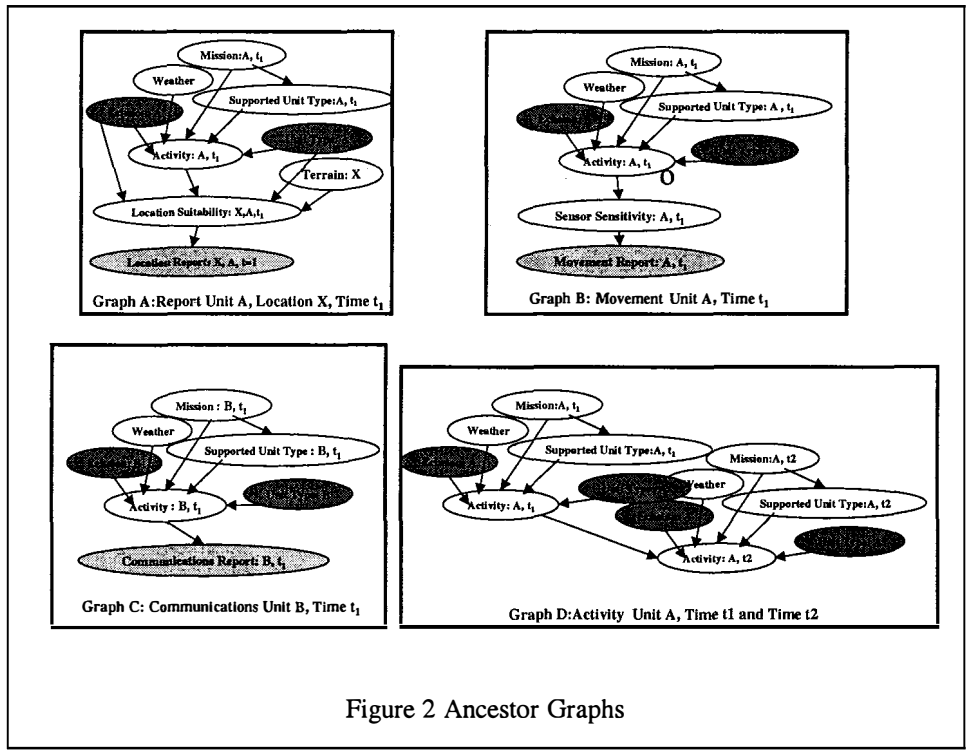

Figure 2 Ancestor Graphs

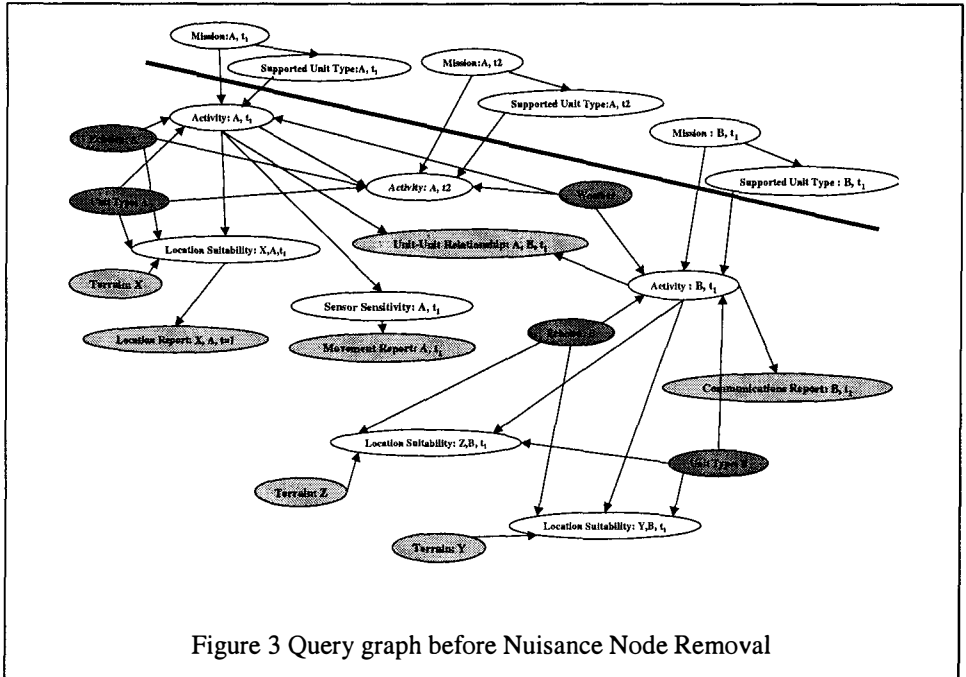

Figure 3 Query graph before Nuisance Node Removal

A network fragment may contain barren nodes that are not in the query graph. These are removed by the construction algorithm.

The network fragment for "Activity" carries several distributions, two of which are represented by the **A1/A2** fragment of Figure 4. Local distribution **A1** with six parents for "Activity" represents its full distribution, while Local Distribution **A2** with just three parents represents the distribution marginalized over three of the parents. The query graph tells the constructor which parents condition each instance of "Activity." The parameters for the *GetLocalDistribution* method now include a list of the parent variables. Given the parent set, the method returns the proper distribution for the node.

Figure 5 shows the constructed situation-specific

*"Unit Type"(Unit A)=Artillery* and *"Echelon"(Unit A)=Battery.* Graph D of Figure 2 shows the ancestor graph for "Activity" with the appropriate identifying attributes. It has two "Activity" vertices and their parents because two times were specified. In this case the method for retrieving ancestor graphs produces a graph in which all the specified identifying attributes are present.

The final predecessor graph is shown in Figure 3. Removing the nuisance nodes using the algorithm of Lin and Druzdzel eliminates all of the "Mission" and "Supported Unit Type" nodes. These are the vertices above the dark line running through the graph.

In the final phase of network construction the constructor retrieves network fragments from the knowledge base, instantiates them, and adds them to the situation-specific network. The query graph guides the retrieval of the network fragments.

For most variables, the network constructor can read the fragment retrieval query from the query graph. The query graph provides the variable names, identifying attributes, marked evidence and context vertices, and contextual dependencies. The query itself provides the values for context

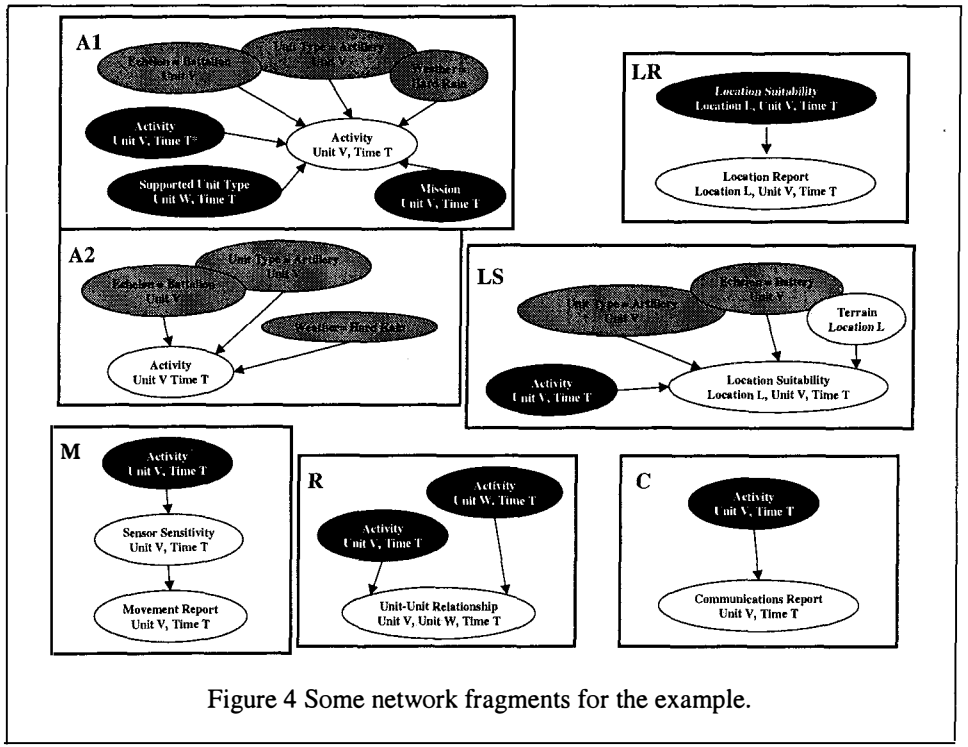

Figure 4 Some network fragments for the example.

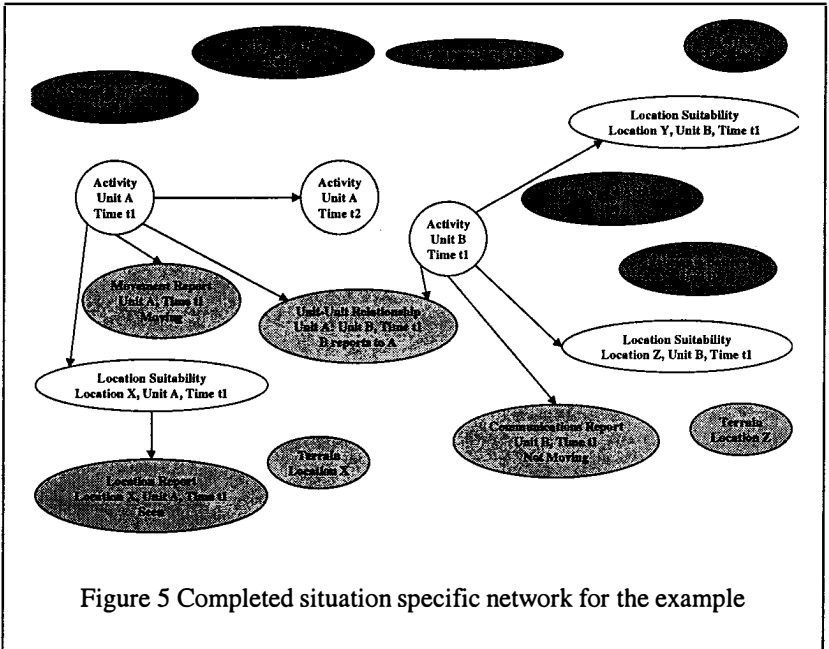

Figure 5 Completed situation specific network for the example

network.

## Acknowledgements

The research reported in this paper was sponsored by DARPA and the U.S. Army Topographic Engineering Center under contract DACA76-93-0025 to Information Extraction and Transport, Inc. The authors thank Steve Langs of IET for many helpful discussions. In addition, we extend grateful acknowledgment to three anonymous reviewers for helpful comments and suggestions on an earlier version of this paper.